\documentclass[runningheads]{llncs}
\usepackage{makeidx}  
\usepackage[colorlinks,linkcolor=blue, urlcolor=blue]{hyperref}
\usepackage{url}
\usepackage{longtable}
\usepackage{multirow}
\usepackage{amssymb, amsmath, bm}
\usepackage{amsfonts, amssymb}
\usepackage{pifont}
\usepackage{mathtools}
\usepackage{mathrsfs}
\usepackage{booktabs}
\usepackage{cite}
\usepackage{dsfont}
\usepackage{makecell}
\usepackage[noend]{algpseudocode}
\usepackage{algorithmicx,algorithm}
\usepackage[misc]{ifsym} 
\usepackage{bbding}

\begin{document}
\title{Robust Box Prompt based SAM for \\ Medical Image Segmentation}

\author{Yuhao Huang\inst{1,2}, Xin Yang\inst{1,2} \and Han Zhou\inst{1,2} \and Yan Cao\inst{3} \and \\Haoran Dou\inst{4} \and Fajin Dong\inst{5} \and Dong Ni\inst{1,2}\textsuperscript{(\Letter)}} 
\authorrunning{Y. Huang et al.}

\institute{
\textsuperscript{$1$}National-Regional Key Technology Engineering Laboratory for Medical Ultrasound, School of Biomedical Engineering, Medical School, Shenzhen University, China\\
\email{nidong@szu.edu.cn} \\
\textsuperscript{$2$}Medical Ultrasound Image Computing (MUSIC) Lab, Shenzhen University, China\\
\textsuperscript{$3$}Shenzhen RayShape Medical Technology Co., Ltd, China\\
\textsuperscript{$4$}Centre for Computational Imaging and Simulation Technologies in Biomedicine (CISTIB), University of Leeds, UK\\
\textsuperscript{$5$}First Affiliated Hospital, Southern University of Science and Technology, Shenzhen People’s Hospital, China}
\maketitle              

\begin{abstract}
The Segment Anything Model (SAM) can achieve satisfactory segmentation performance under high-quality box prompts.
However, SAM's robustness is compromised by the decline in box quality, limiting its practicality in clinical reality.
In this study, we propose a novel \textbf{Ro}bust \textbf{Box} prompt based SAM (\textbf{RoBox-SAM}) to ensure SAM's segmentation performance under prompts with different qualities.
Our contribution is three-fold.
First, we propose a prompt refinement module to implicitly perceive the potential targets, and output the offsets to directly transform the low-quality box prompt into a high-quality one.
We then provide an online iterative strategy for further prompt refinement.
Second, we introduce a prompt enhancement module to automatically generate point prompts to assist the box-promptable segmentation effectively.
Last, we build a self-information extractor to encode the prior information from the input image.
These features can optimize the image embeddings and attention calculation, thus, the robustness of SAM can be further enhanced.
Extensive experiments on the large medical segmentation dataset including 99,299 images, 5 modalities, and 25 organs/targets validated the efficacy of our proposed RoBox-SAM.
\end{abstract}

\section{Introduction}
\label{sec:intro}
The Segment Anything Model (SAM) has marked a significant breakthrough in image segmentation, due to its powerful zero-shot performance~\cite {Kirillov2023segment}.
SAM was trained on a large annotated dataset with 11M images and 1B masks based on the Vision in Transformer (ViT).
It supports different types of prompts, including point, box, text, and mask. 
Recent studies have widely explored the potential of applying SAM to the medical imaging domain~\cite{mazurowski2023segment,huang2024segment}, which demonstrates that SAM can outperform the task-specific state-of-the-art conventional deep models after fine-tunning with additional medical images~\cite {ma2024segment}.

\begin{figure*}[!t]
	\centering
	\includegraphics[width=0.95\linewidth]{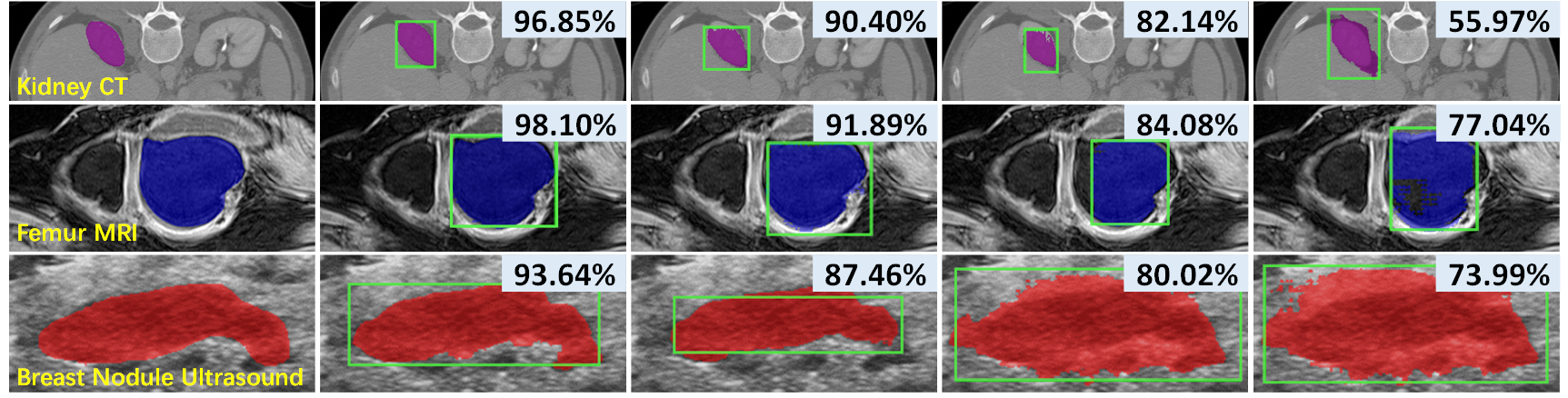}
	\caption{
 Performance of SAM under box prompts. 
 Columns 1-2: annotated masks and the SAM's performance under tight box prompts. 
 Columns 3-5: performance under different low-quality prompts. DICE metrics are shown in the right-up corners.}
	\label{fig:intro}
\end{figure*}

Most previous studies equipped the SAM with high-quality prompts, e.g., the center point or tight box of the target.
However, this assumption typically cannot be held in clinical practice, where numerous low-quality prompts commonly occur.
As shown in Fig.~\ref{fig:intro}, SAM prompted with high-quality boxes can achieve good performance, whereas it will easily generate poor segmentation after randomly shifting the promoting boxes. 
Hence, enhancing the robustness of SAM against imprecise prompts is crucial for its reliability.

\textbf{Segmentation ability improvement of SAM.}
Fine-tuning SAM using medical datasets to enhance medical image segmentation efficacy is an important field of research. 
Ma \textit{et al.} ﬁxed the prompt encoder, and tuned the parameters in both image encoder and mask decoder during training~\cite{ma2024segment}.
Wu \textit{et al.} then introduced the adaptor modules with few trainable parameters to capture medical knowledge~\cite{wu2023medical}.
Xie \textit{et al.} reconstructed the mask decoder within SAM to develop a few-shot fine-tuning strategy~\cite{xie2024sam}.
Moreover, Semantic-SAM~\cite{li2023semantic} and HQ-SAM\cite{ke2024segment} were proposed to improve the segmentation quality. 
However, these studies aimed to improve the segmentation performance of SAM in different scenarios. 
They assumed to directly use the high-quality tight box as prompt, lacking the exploration of the low-quality prompt.

\textbf{Robustness of SAM.} 
Several studies have evaluated SAM's robustness under various corruptions and adversarial attacks~\cite{huang2023robustness,wang2023empirical,qiao2023robustness}. 
However, their robustness tests were conducted at the image level rather than the prompt level. 
In~\cite{huang2024segment}, the authors randomly shifted the tight box to mimic the practical operation.
They concluded that the SAM is sensitive to prompt randomness, especially box prompts.
The higher the random shifts, the more serious the performance degradation.
Although they have noticed the weak robustness of SAM under different prompt shifts, they did not introduce solutions to solve this problem.

One study observed that when subjected to low-quality prompts, SAM's mask decoder encounters an attention drift issue, leading to the cross-attention module activating inaccurate regions~\cite{fan2023stable}.
To address this issue, stable SAM (SSAM) leveraged a deformable sampling plugin to shift attention to the target region.
However, SSAM only predicted the indirect feature offset to enhance the robustness.
Besides, it was developed and validated on natural images characterized by clear target boundaries, distinguishable foregrounds and backgrounds, and color encoding. 
These characteristics significantly differ from those of medical images, which may present unique challenges to the SSAM.

In this study, we introduce a novel robust box prompt based SAM named RoBox-SAM, aimed at making SAM robust to various prompt qualities.
Our contribution is three-fold.
First, we introduce an intuitive prompt refinement module (PRM) designed for potential target perception and directly predict box prompt offsets to refine low-quality prompts.
We also design an online refinement strategy to optimize the box prompt quality iteratively, and further achieve robustness improvement.
Second, we propose the prompt enhancement module (PEM), and leverage the movement information between the noisy and optimized prompts to output the potential point prompts automatically.
Finally, we extract intrinsic features from the input image via a self-information extractor (SIE), to enrich the image embeddings and refine the cross-attention calculation within the decoder. 
Validated on the large medical segmentation dataset, we demonstrate that RoBox-SAM is robust to box prompts of varying qualities, and general for different medical modalities and targets, achieving satisfactory results.

\section{Methodology}
\label{sec:Method}
\begin{figure*}[!t]
	\centering
	\includegraphics[width=1.0\linewidth]{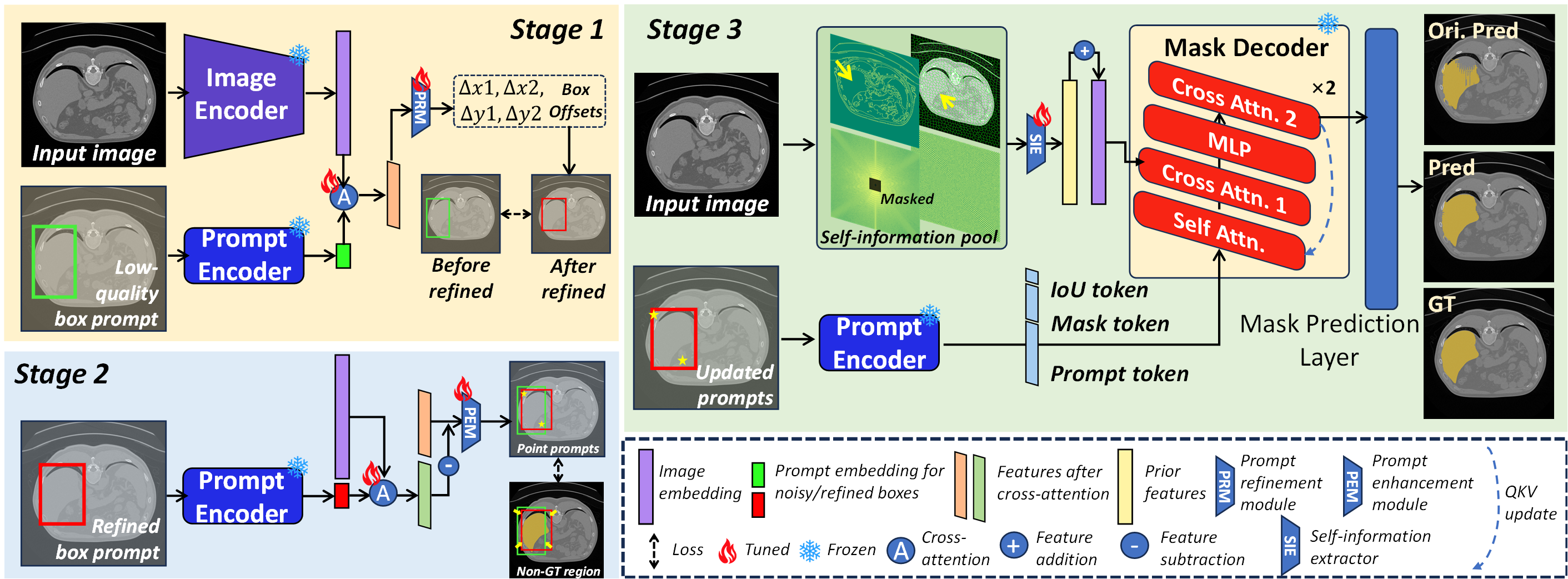}
	\caption{Overview of our proposed framework.}
	\label{fig:framework}
\end{figure*}

Fig.~\ref{fig:framework} shows the schematic view of our proposed Robox-SAM.
In \textit{stage 1}, to eliminate the effects of low-quality prompts, the proposed PRM first predicts the offsets to transfer the input box prompt into a high-quality one.
In \textit{stage 2}, we adopt the prompt correction knowledge to output potential points via the proposed PEM.
In the last stage, we leverage the prior knowledge encoded by SIE to optimize the image embeddings, and update the attention calculation in the mask decoder.
This can further optimize the feature activations under suboptimal box prompts, thus, improving the robustness of SAM.

\textbf{Original SAM.}
SAM mainly consists of three parts: image encoder, prompt encoder, and mask decoder~\cite{Kirillov2023segment}.
The image encoder is used to extract the image embedding, while the prompts are encoded using different methods (e.g., point and box: positional encoding~\cite{tancik2020fourier}, text: CLIP~\cite{radford2021learning}). 
Before being fed to the mask decoder, the image embedding should add the positional encoding, while the prompt tokens should concatenate with two output tokens, i.e., IoU token and mask token, to form the \textit{embedding} ($e$) and \textit{tokens} ($t$), respectively.
The mask decoder mainly contains two \textit{Transformer} layers, and each layer contains four modules: 1) self-attention of \textit{tokens}, 2) token-to-image cross-attention, 3) Multi-Layer Perception (MLP) on \textit{tokens}, and 4) image-to-token cross-attention.
For example, the token-to-image cross-attention can be formulated as $C_{Att}(t,e)$=$softmax(Q(t),K(e)^{T}/\sqrt{d_{k}})\cdot V(e)$, where $d(k)$ represents the dimension of the keys.
Finally, the mask token is separated from \textit{tokens} and operated with the final up-scaled image embedding to predict multiple outputs.
Besides, the IoU token predicts the IoU score for each mask.

\textbf{PRM for Target Perception and Box Prompt Correction.}
Inaccurate box prompts may cause the SAM's mask decoder to activate imprecise regions, and the cross-attention between image features and \textit{tokens} may further accumulate the errors, resulting in unstable segmentation~\cite{fan2023stable}.
We propose the PRM to learn and perceive the target location, and predict the offsets to update the input prompts.
Given the image and prompt features ($f_{img}$, $f_{p}$), cross-attention can be performed to obtain $f^{*}$=$f_{img}+C_{Att}(f_{img},f_{p})$.
Then, $f^{*}$ will be inputted to the light-weighted PRM, outputting the offsets $\Delta(o)$=$\left\{\Delta x_1, \Delta x_2, \Delta y_1, \Delta y_2\right\}$.
The input box prompt ($p_{c}=\left\{x_1,x_2,y_1,y_2\right\}$) can be updated to $p_{c}^{*}$=$p_{c}$+$\Delta(o)$.
The loss function can be written as $L_{o}$=$L_{1}(p_{c}^{*}, p_{gt})$, where $p_{gt}$ represents the coordinate of GT tight box.
This supervision can drive the development of SAM's target perception ability, outputting reliable offsets to make the input box prompt move and match the potential target.
Moreover, during testing, the box refinement process can be interactive. Specifically, the output refined box in the current step can be considered as the input of the next step.
The iteration will terminate when $\sum\Delta(o)$=$0$ or iterating over $k$=5 steps.
This interactive refinement strategy can further improve the model's accuracy and robustness.

\textbf{PEM for Automatic Point Prompt Generation.}
We only consider the box as the initial input prompt. 
However, we believe that the point prompts can assist the box prompts in some cases, especially negative points for restraining over-segmentation.
The above box correction process is intuitive and effective in revealing the potential region for point prompt generation.
Refer to \textit{stage 2} (right-down corner) in Fig.~\ref{fig:framework}, given the original (green, \textit{A}) and refined (red, \textit{B}) boxes, and the annotated ground truth (GT) mask (orange, \textit{C}), the non-GT region (i.e., $(A\cup B)$-$C$) indicated by yellow arrows can be treated as the potential negative area.
This region can be flattened into a one-dimensional vector and \textit{n} uniformly sampled points can be taken as GTs ($p_{gt}$).
The CNN-based PEM ($\mathcal{M}$) can learn the potential points by inputting image-prompt attention features before and after refinement ($p_{pre}$=$\mathcal{M}$($f_{r}^{*}$-$f^{*}$)).
Specifically, it generates \textit{n=5} points and matches them with sampled GT points one by one to find the combination with the smallest distance errors to calculate the L2 loss: $L_{p}$=$||(p_{pre},p_{gt})||_{2}$

\textbf{Self-information Mining for Feature and Attention Optimization.}
Utilizing the prior knowledge of the input image can enhance the accuracy and robustness of the segmentation model.
In the self-information pool, we provide several components that contain useful shape and edge information.
As shown in the green block of Fig.~\ref{fig:framework} (\textit{stage 3}), yellow arrows in the canny-based sketch~\cite{canny1986computational} and SLIC-based superpixel~\cite{achanta2012slic} represent the location and boundary of the target.
We also adopt the fast Fourier Transform to transfer the image to the frequency domain~\cite{yang2020fda,huang2023fourier}, including amplitude and phase parts.
Specifically, the low-frequency amplitudes are masked to highlight the details and edges of the image.
This information was encoded by the proposed SIE, and the output dimension was mapped to the same as the image embedding.
Then, the image embedding can be re-calculated by $f'$=$f^{*}$+$f_s$ to fuse and emphasize the self-information.
Finally, $f'$ will be input into the cross-attention module for attention re-calculation and optimization.
Thus, though the refined box prompts may still include noise, the inherent self-information can help remove it via attention update, and assist in improvement in terms of both accuracy and robustness.

\section{Experimental Results}
\label{sec:Experimental Results}
\textbf{Materials and Implementations.} 
We validated the RoBox-SAM on a large medical segmentation dataset, including 99,299 images/slices, 5 modalities, and 25 types of targets.
This dataset contains 14 public and 5 in-house datasets (marked with $^*$).
Specifically, for \textbf{CT} datasets~\cite{ma2021abdomenct,ji2022amos,chestsegmentation}, the targets include chest (heart, lung, trachea) and abdomen (aorta, bladder, gall bladder, kidney, liver, pancreas, spleen, stomach).
\textbf{MRI} datasets contain the brain~\cite{avital2019neural}, brain tumor~\cite{ji2021learning}, knee (femur and tibia)~\cite{lee2010learning}, and prostate~\cite{lemaitre2015computer}.
\textbf{Ultrasound} datasets have nine objects, including breast nodule$^*$, eyes$^*$, femur$^*$, heart~\cite{leclerc2019deep} (left atrium, left ventricle endocardium, left ventricle epicardium), kidney pelvis$^*$, nuchal translucency$^*$, and thyroid nodule~\cite{thyroidsegmentation}.
We also considered two other targets from two modalities, i.e., \textbf{Colonoscopy}: polyp~\cite{hicks2021endotect} and \textbf{Fundus}: OpticDisc~\cite{huazhu2019palm,li2020self}.
Approved by the local IRB, the in-house ultrasound datasets were collected and annotated by an experienced sonographer under strict quality control.
CT/MR volumes were sliced into 2D images similar to~\cite{huang2024segment}.
The whole dataset was split into 6:1:3 for fine-tuning, validation, and testing, respectively.

We implemented RoBox-SAM in Pytorch, using an NVIDIA A40 GPU.
The batch size, learning rate, and total epoch are set to 16, 1e-4, and 20, respectively.
SAM's backbone was \textit{ViT-H}.
GT box was generated by the adjacency box of the annotated mask.
Input box prompts were randomly shifted by 0-30\% based on the GT box to evenly mimic the human operations with varying experiences/annotation preferences, with the condition that their IoU$>$0.5.
We followed the SAM to use cross-entropy and dice losses to optimize the model via AdamW optimizer.
We further added $L_{o}$ and $L_{p}$ to assist the model learning.
The proposed PRM, PEM, and SIE are two-layer CNNs with linear projection.
They encode the high-dimensional features and output offsets, points, or low-dimensional features to satisfy specific requirements.
As shown in Fig.~\ref{fig:framework}, during fine-tuning, the original components in SAM were frozen, whereas other parts were learnable.
During testing, all the parameters were frozen, and the mask decoder will output three predictions with corresponding predicted IoU scores.
The mask with the highest IoU prediction was selected as the final output.

\begin{table}[!h]
  \centering
  \scriptsize
  \caption{Method comparison under low-quality box prompt with 10\% random shift. The best results are shown in bold.}
      \resizebox{\textwidth}{!}{
    \begin{tabular}{c|cc|cc|cc|cc|cc}
    \toprule
          & \multicolumn{2}{c|}{MRI} & \multicolumn{2}{c|}{CT} & \multicolumn{2}{c|}{Ultrasound} & \multicolumn{2}{c|}{Colonoscopy} & \multicolumn{2}{c}{Fundus} \\
    \midrule
    Methods & DICE  & PR   & DICE  & PR   & DICE  & PR   & DICE  & PR   & DICE  & PR \\
     \midrule
    SAM$_{ViT-B}$   & 72.10 & 68.67 & 73.82 & 65.24 & 63.17 & 58.70 & 75.08 & 64.71 & 72.23 & 66.17 \\
    SAM$_{ViT-L}$   & 74.12 & 69.41 & 75.27 & 66.40 & 66.22 & 62.50 & 77.23 & 69.18 & 76.45 & 71.77 \\
    SAM$_{ViT-H}$   & 75.36 & 68.58 & 74.82 & 68.77 & 70.63 & 64.41 & 80.96 & 75.42 & 81.23 & 78.54 \\
    MedSAM (Baseline) & 78.68 & 70.25 & 77.81 & 70.15 & 73.48 & 67.19 & 83.34 & 77.98 & 82.17 & 79.96 \\
    MedSAM$_{boxaug}$ & 80.18 & 71.72 & 79.49 & 71.83 & 75.23 & 69.77 & 86.77 & 80.08 & 85.43 & 85.52 \\
    MedSAM$_{offset}$ & 76.45 & 66.21 & 77.04 & 68.03 & 71.68 & 65.56 & 84.98 & 76.22 & 83.21 & 80.09 \\
    HQ-SAM & 81.22 & 76.34 & 80.11 & 72.31 & 73.84 & 67.48 & 86.22 & 79.17 & 85.47 & 82.74\\
    \midrule
    Baseline+PRM & 84.97 & 80.08 & 83.93 & 78.44 & 79.21 & 73.24 & 89.21 & 87.89 & 87.11 & 86.31 \\
    Baseline+PRM+PEM & 87.72 & 84.36 & 86.88 & 83.09 & 81.33 & 76.13 & 91.34 & 90.02 & 89.23 & 87.66 \\
    RoBox-SAM  & \textbf{90.09} & \textbf{87.89} & \textbf{88.02} & \textbf{87.17} & \textbf{83.17} & \textbf{80.78} & \textbf{93.77} & \textbf{91.89} & \textbf{91.72} & \textbf{89.94} \\
    \bottomrule
    \end{tabular}}
  \label{tab:1}%
\end{table}%

\begin{table}[!t]
  \centering
  \scriptsize
  \caption{Method comparison under low-quality box prompt with 0-30\% random shift.}
  \setlength{\tabcolsep}{0.9mm}
    \begin{tabular}{cc|cc|cc|cc|cc|cc}
    \toprule
          &       & \multicolumn{2}{c|}{MRI} & \multicolumn{2}{c|}{CT} & \multicolumn{2}{c|}{Ultrasound} & \multicolumn{2}{c|}{Colonoscopy} & \multicolumn{2}{c}{Fundus} \\
    \midrule
          & Methods & DICE  & PR   & DICE  & PR   & DICE  & PR   & DICE  & PR   & DICE  & PR \\
    \midrule
    \multirow{2}[0]{*}{GT} & Med-SAM & 90.83  & /     & 89.06  & /     & 84.78  & /     & 92.24  & /     & 92.77  & / \\
          & RoBox-SAM  & 91.07  & /     & 89.45  & /     & 85.02  & /     & 93.88  & /     & 93.18  & / \\
    \midrule
    \multirow{3}[0]{*}{0-10\%} & Med-SAM & 78.68  & 70.25  & 77.81  & 70.15  & 73.48  & 67.19  & 83.34  & 77.98  & 82.17  & 79.96  \\
          & RoBox-SAM  & 90.09  & 87.89  & 88.02  & 87.17  & 83.17  & 80.78  & 93.77  & 91.89  & 91.72  & 89.94  \\
          & RoBox-SAM (Iter) & 90.13  & 88.31  & 88.49  & 87.43  & 83.26  & 81.00  & 93.61  & 91.92  & 91.86  & 90.04  \\
    \midrule
    \multirow{3}[0]{*}{10-20\%} & Med-SAM & 65.22  & 59.11  & 65.83  & 59.43  & 63.15  & 55.10  & 76.72  & 69.03  & 75.34  & 67.33  \\
          & RoBox-SAM  & 88.41  & 86.47  & 86.33  & 83.08  & 81.77  & 79.01  & 92.08  & 90.79  & 89.21  & 88.05  \\
          & RoBox-SAM (Iter) & 89.71  & 87.23  & 87.14  & 85.02  & 83.01  & 80.64  & 92.43  & 91.08  & 90.74  & 89.15  \\
    \midrule
    \multirow{3}[0]{*}{20-30\%} & Med-SAM & 53.33  & 49.20  & 49.98  & 43.84  & 53.24  & 48.39  & 68.53  & 62.27  & 62.14  & 57.56  \\
          & RoBox-SAM  & 85.41  & 83.05  & 84.19  & 81.69  & 79.33  & 77.74  & 90.67  & 88.36  & 87.36  & 85.42  \\
          & RoBox-SAM (Iter) & 86.63  & 83.93  & 85.24  & 82.73  & 80.10  & 78.90  & 91.74  & 89.32  & 88.04  & 86.79  \\
    \bottomrule
    \end{tabular}%
  \label{tab:2}%
\end{table}%

\textbf{Quantitative and Qualitative Analysis.}
We performed \textit{N=5} times random testing, and reported the average performance to obtain stable and reliable results.
Evaluation metrics in terms of accuracy (DICE, \%) and robustness to prompt (PR, \%: $\sum_{i=1}^{N}(DICE(M_i, M_u))/N$, similar to~\cite{fan2023stable}), were used to validate the segmentation performance.
$M_i$ represents the mask in $i$-th testing, while $M_u$ is the union mask region of all $N$ masks.
The higher the PR score, the more consistent the segmentation results are across box prompts of different qualities.

Table~\ref{tab:1} compares the RoBox-SAM with 7 methods, including SAM~\cite{Kirillov2023segment} (different model sizes: \textit{ViT-B/L/H}), MedSAM~\cite{ma2024segment}, MedSAM$_{boxaug}$, MedSAM$_{offset}$, and HQ-SAM~\cite{ke2024segment}. 
All the MedSAM-based methods and HQ-SAM are based on \textit{ViT-H} and fine-tuned using our medical datasets.
MedSAM$_{boxaug}$ adds the random box augmentation strategy during fine-tuning, and MedSAM$_{offset}$ trains an additional detector to predict the offsets for prompt correction before inputting to SAM. 
Based on the first five experiments, we concluded that enlarging the model size, fine-tuning using medical images, and augmenting with random boxes can optimize model robustness in most cases.
However, comparing MedSAM with MedSAM$_{offset}$, we found that adding naive offset prediction cannot steadily improve robustness, with 3/5 modalities having performance drops.
RoBox-SAM achieves the best performance among all the competitors (\textit{p$<$0.05}), including the strong HQ-SAM, showing its efficacy and superiority in model robustness improvement.
Besides, RoBox-SAM is economical, has low training learnable computational overhead ($<$1M params), and takes only $\sim$0.6s per testing image.

\begin{figure*}[!t]
	\centering
	\includegraphics[width=1.0\linewidth]{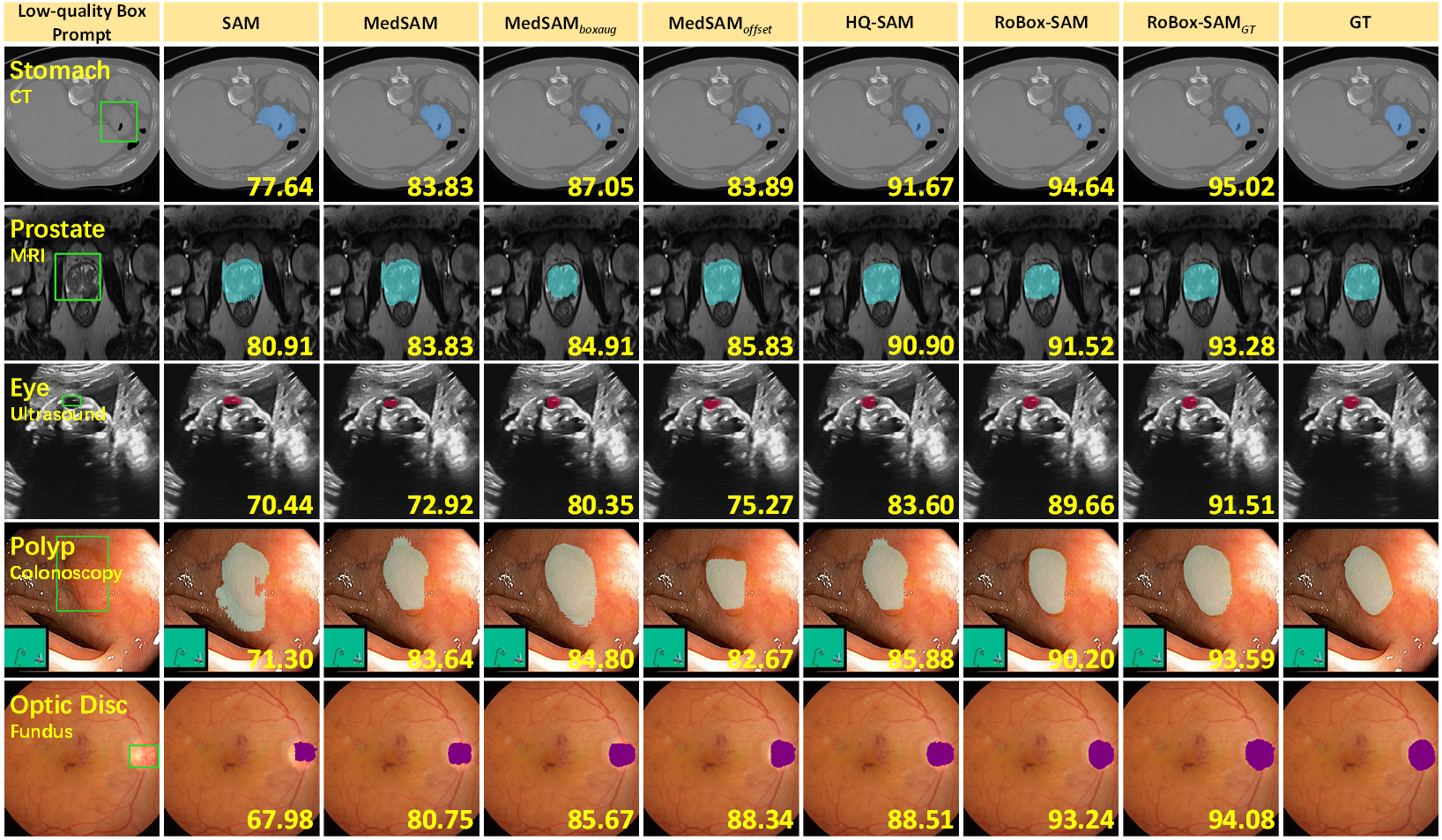}
	\caption{Visualization results of different methods.}
	\label{fig:result1}
\end{figure*}

We also performed ablation studies to evaluate the contribution of each proposed module in Table~\ref{tab:1}.
Specifically, we treated MedSAM as \textit{baseline}, and progressively added the proposed modules to observe the performance changes.
It can be seen that PRM contributes $\sim$6\% and $\sim$8\% improvements to DICE and PR scores, respectively.
Adding PEM, the metrics can be increased further.
This illustrates that the generated point prompts can effectively assist box prompts in achieving better segmentation. 
By aggregating the prior self-information (RoBox-SAM), the model's performance can achieve consistent improvement across the five modalities.
It is highlighted that the performance improvement brought by each module is statistically significant with \textit{p$<$0.05}.

\begin{figure*}[!t]
	\centering
	\includegraphics[width=1.0\linewidth]{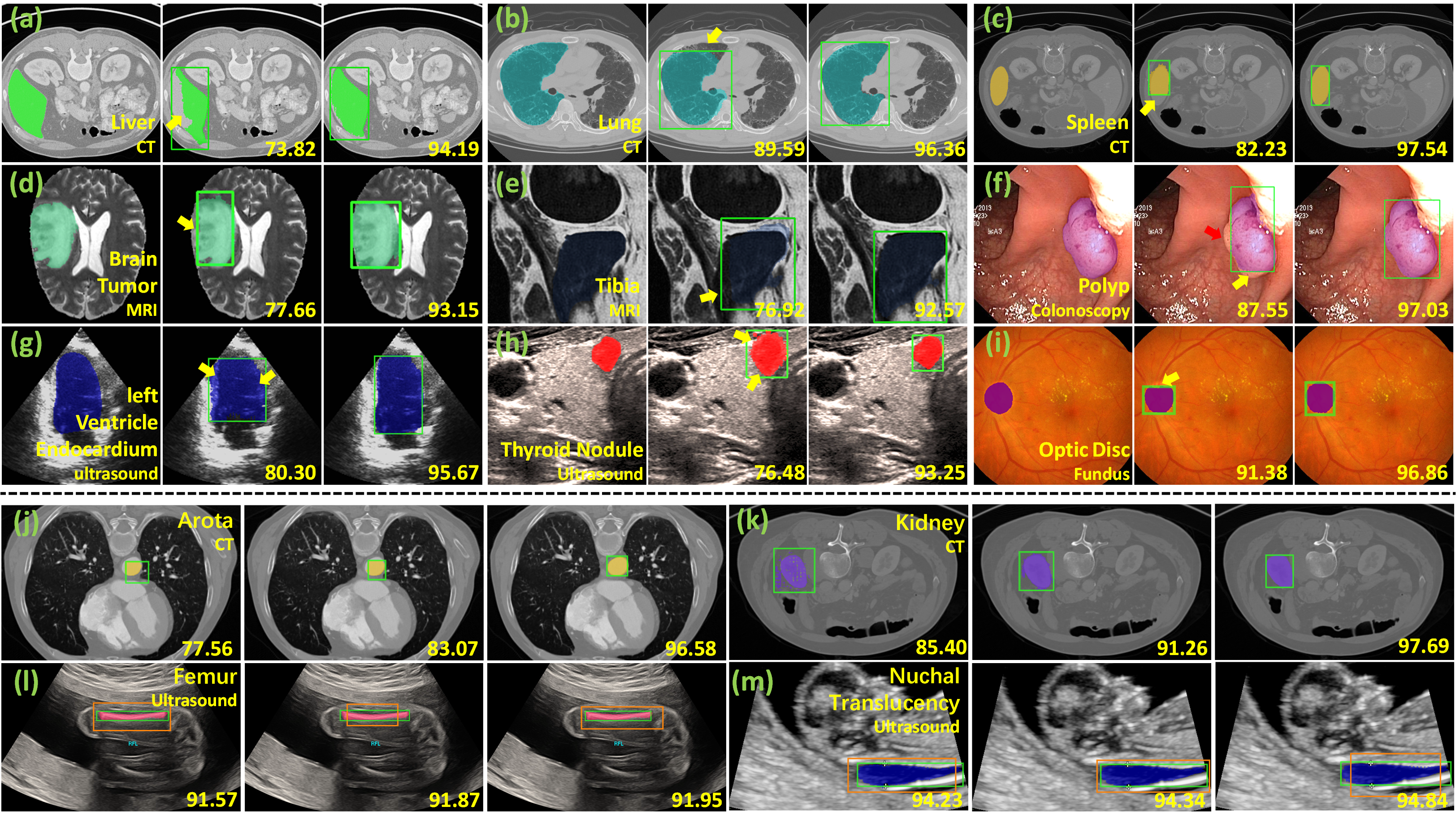}
	\caption{Visualization of typical cases. Yellow arrows show the segmentation errors.} 
	\label{fig:result2}
\end{figure*}

In Table~\ref{tab:2}, we further decreased the quality of box prompts by progressively increasing the random shift max to 30\%.
This can verify the model's performance under extreme conditions.
The lines \textit{GT} mean that the tight boxes of objects are taken as input box prompts.
In this situation, although the PRM may not output large offsets, point prompt enhancement and self-information mining can still improve the segmentation.
Thus, RoBox-SAM shows a slight improvement compared with MedSAM under high-quality tight box prompts.
As the box prompt quality continued to decline, the accuracy and stability of MedSAM severely degraded.
For example, CT targets dropped about 50\% in DICE values (89.06\%$\rightarrow$49.98\%).
The performance of RoBox-SAM was less affected by low-quality box prompts (e.g., CT: 5.26\%$\downarrow$ in DICE).
This shows that RoBox-SAM can cope well with extreme box prompts and output more stable results than MedSAM.
Moreover, we provided the results of RoBox-SAM (Iter), and found that the interaction strategy can help boost the DICE and PR scores, especially for extremely low-quality box prompts ($>1\%$ significant improvement).

Fig.~\ref{fig:result1} and Fig.~\ref{fig:result2} are qualitative results, with DICE scores shown in the right-down corners of the images.
Fig.~\ref{fig:result1} visualizes the segmentation results of compared methods and our proposed RoBox-SAM.
It can be observed that the original SAM shows poor performance under low-quality prompts.
RoBox-SAM performs better than other methods, approaching the upper-bound (RoBox-SAM$_{GT}$) and GT.
Fig.~\ref{fig:result2} (a)-(i) shows the box refinement and performance comparison (in each sub-figure, from left to right shows GT, MedSAM, and RoBox-SAM results).
Given low-quality box prompts, MedSAM produces poor prediction, while RoBox-SAM can refine the box and output satisfactory results.
Fig.~\ref{fig:result2} (j)-(k) shows the interactive prompt refinement process.
RoBox-SAM performs worse in the first-round refinement, but achieves good correction finally due to the iterative refinement strategy.
Fig.~\ref{fig:result2} (l)-(m) reveals the robustness of RoBox-SAM. In each case, given three different random prompts (orange boxes), RoBox-SAM can refine them to the green ones, and output stable results.

\section{Conclusion}
\label{sec:conclusion}
In this paper, we propose RoBox-SAM for improving SAM's robustness to box prompts of different qualities.
We first directly refine the input box with poor quality via the proposed PRM to obtain a high-quality box prompt.
Then, we automatically generate potential point prompts through REM to assist the box-promptable segmentation.
Last, we mine the self-information in the image to further ensure robustness via attention optimization.
Extensive experiments on the large medical dataset validate the efficacy of RoBox-SAM.
In the future, we will extend the framework to more modalities and 3D segmentation tasks.

\begin{credits}
\subsubsection{\ackname}
This work was supported by the grant from National Natural Science Foundation of China (12326619, 62101343, 62171290), Science and Technology Planning Project of Guangdong Province (2023A0505020002), and Shenzhen-Hong Kong Joint Research Program (SGDX20201103095613036).

\subsubsection{\discintname}
The authors have no competing interests to declare that are relevant to the content of this article.
\end{credits}

\bibliographystyle{splncs04}
\bibliography{reference}

\end{document}